\documentclass{article}
\usepackage{arxiv}
\pdfoutput=1

\usepackage{amsmath,amsfonts,bm}

\newcommand{\RR}{\mathbb{R}} 
\def\eqref#1{Equation~(\ref{#1})}

\def\1{\bm{1}}

\DeclareMathAlphabet{\mathsfit}{\encodingdefault}{\sfdefault}{m}{sl}
\SetMathAlphabet{\mathsfit}{bold}{\encodingdefault}{\sfdefault}{bx}{n}

\def\gD{{\mathcal{D}}}
\def\gE{{\mathcal{E}}}

\def\gL{{\mathcal{L}}}

\def\gX{{\mathcal{X}}}
\def\gY{{\mathcal{Y}}}

\usepackage{amsthm}
\usepackage{amsmath}
\usepackage{multirow, multicol}
\usepackage{amsfonts}
\usepackage{tabularx}
\usepackage{graphicx}
\usepackage{bm}
\usepackage{hyperref}
\usepackage{diagbox}
\usepackage{threeparttable}
\usepackage{subcaption}
\usepackage{algorithm}
\usepackage{algpseudocode}
\usepackage{xspace}
\usepackage{xcolor}
\usepackage{enumitem}
\usepackage[numbers]{natbib}
\usepackage{url}
\usepackage{booktabs}

\newcommand{\xmc}{XMC\xspace}

\newcommand{\gzxmc}{GZ-XMC\xspace}
\newcommand{\setting}{EZ-XMC\xspace}
\newcommand{\fsxmc}{FS-XMC\xspace}
\newcommand{\amznsmall}{LF-Amazon-131K\xspace}
\newcommand{\amznlarge}{LF-Amazon-1M\xspace}
\newcommand{\wikismall}{LF-WikiSeeAlso-320K\xspace}
\newcommand{\wikilarge}{LF-Wikipedia-500K\xspace}
\newcommand{\tfidf}{TF-IDF\xspace}
\newcommand{\xrl}{XR-Linear\xspace}
\newcommand{\glove}{GloVe\xspace}
\newcommand{\sbert}{SentBERT\xspace}
\newcommand{\mpnet}{MPNet\xspace}
\newcommand{\ict}{ICT\xspace}
\newcommand{\simcse}{SimCSE\xspace}
\newcommand{\astec}{Astec\xspace}
\newcommand{\attnxml}{AttentionXML\xspace}
\newcommand{\siamesexml}{SiameseXML\xspace}
\newcommand{\xrtransformer}{XR-Transformer\xspace}
\newcommand{\zestxml}{ZestXML\xspace}
\newcommand{\ours}{MACLR\xspace}

\algnewcommand{\LeftComment}[1]{\Statex \(\triangleright\) #1}

\title{Extreme Zero-Shot Learning for Extreme Text Classification}
\date{}

\author{
    {Yuanhao Xiong} \\
    \And
    {Wei-Cheng Chang} \\
    \And
    {Cho-Jui Hsieh} \\
    \And
    {Hsiang-Fu Yu} \\
    \And
    {Inderjit Dhillon}
}

\begin{document}
\maketitle

\begin{abstract}
The eXtreme Multi-label text Classification (\xmc) problem concerns
	finding most relevant labels for an input text instance from a large label set.
Many real-world applications can be formulated as \xmc problems,
	such as recommendation systems, document tagging and semantic search.
However, the \xmc setup faces two challenges:
	(1) it is not generalizable to predict unseen labels in dynamic environments,
	and (2) it requires a large amount of supervised (instance, label) pairs,
	which can be difficult to obtain for emerging domains.
Recently, the generalized zero-shot \xmc (\gzxmc) setup has been studied
	and \zestxml is proposed accordingly to handle the unseen labels, which still requires a large number of annotated (instance, label) pairs.
In this paper, we consider a more practical scenario called Extreme Zero-Shot \xmc (\setting),
	in which no supervision is needed and merely raw text of instances and labels are accessible.
  Few-Shot XMC~(\fsxmc), an extension to \setting with limited supervision is also investigated.
To learn the semantic embeddings of instances and labels with raw text,
	we propose to pre-train Transformer-based encoders with self-supervised contrastive losses.
Specifically, we develop a pre-training method \textbf{\ours}, which thoroughly leverages the raw text with techniques including
\textbf{M}ulti-scale \textbf{A}daptive \textbf{C}lustering, \textbf{L}abel \textbf{R}egularization, and self-training with pseudo positive pairs.
Experimental results on four public \setting datasets demonstrate that
	\ours achieves superior performance compared to all other leading baseline methods,
  in particular with approximately 5-10\% improvement in precision and recall on average.
  Moreover, we also show that our pre-trained encoder can be further improved on \fsxmc
	when there are a limited number of ground-truth positive pairs in training.
By fine-tuning the encoder on such a few-shot subset,
	\ours still outperforms other extreme classifiers significantly. Our code is available at \url{https://github.com/amzn/pecos/tree/mainline/examples/MACLR}. 
\end{abstract}


\maketitle
\bibliographystyle{plain}
\section{Introduction}
The e\textbf{X}treme \textbf{M}ulti-label text \textbf{C}lassification (\xmc) problem
	aims at tagging a text input with most relevant subset of labels from an extremely large output space.
Many web-related applications can be formulated as an \xmc task with encouraging results, such as
finding the best matching products from a large catalog in e-commerce systems~\citep{medini2019extreme,chang2021extreme},
auto-completing queries given its prefix on search engines~\citep{yadav2021session},
predicting search keywords for dynamic advertising~\citep{prabhu2018parabel,chang2020taming},
tagging categories of Wikipedia articles from a large label taxonomy~\citep{dekel2010multiclass,chalkidis2019large},
to name just a few.

\begin{figure}[t]
  \centering
  \includegraphics[width=0.85\textwidth,bb=0 0 4486 1792]{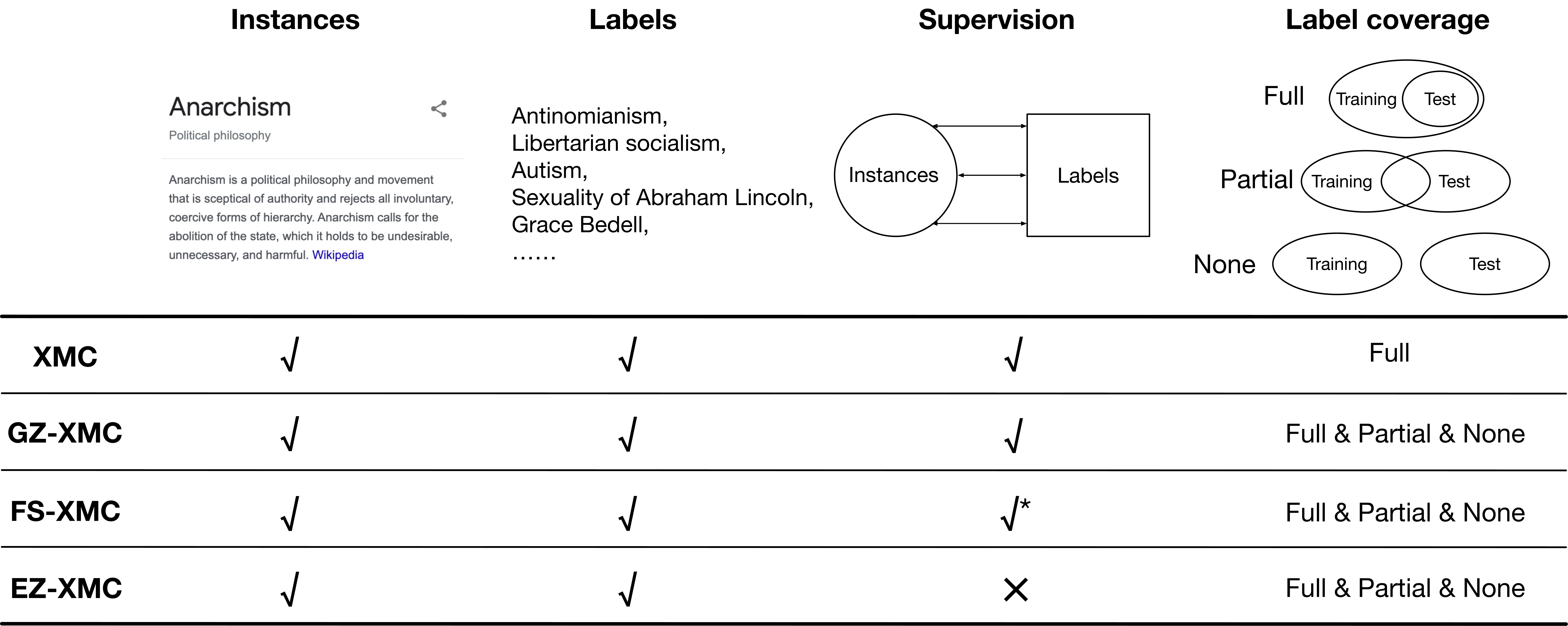}
  \caption{Four different settings in XMC.
  Four essential components are considered: instances~(raw text), labels~(raw text), supervision~(positive pairs), and label coverage.
  In detail, we divide label coverage into 3 groups: full, partial, and none.
  * in \fsxmc emphasizes that only a limited amount of supervision is available.
  We can see that \setting is the most general and practical setting, where no supervision and label coverage is required.}
  \vspace{-1em}
  \label{fig:scenarios}
\end{figure}

The current \xmc setup is built on full label coverage and full supervision, where full label coverage means labels to be predicted have already appeared in the training set and full supervision indicates it requires a significant number of annotated (instance, label) pairs.
In detail, it is assumed that an \xmc algorithm has access to raw text of instances and labels, together with their corresponding relations during training, as shown in Figure \ref{fig:scenarios}.

\begin{figure}[ht]
  \centering
  \begin{subfigure}[ht]{0.4\textwidth}
    \centering
    \includegraphics[width=\textwidth,
    trim={0in 0in 0in 0in},
    clip=false,bb=0 0 394 268]{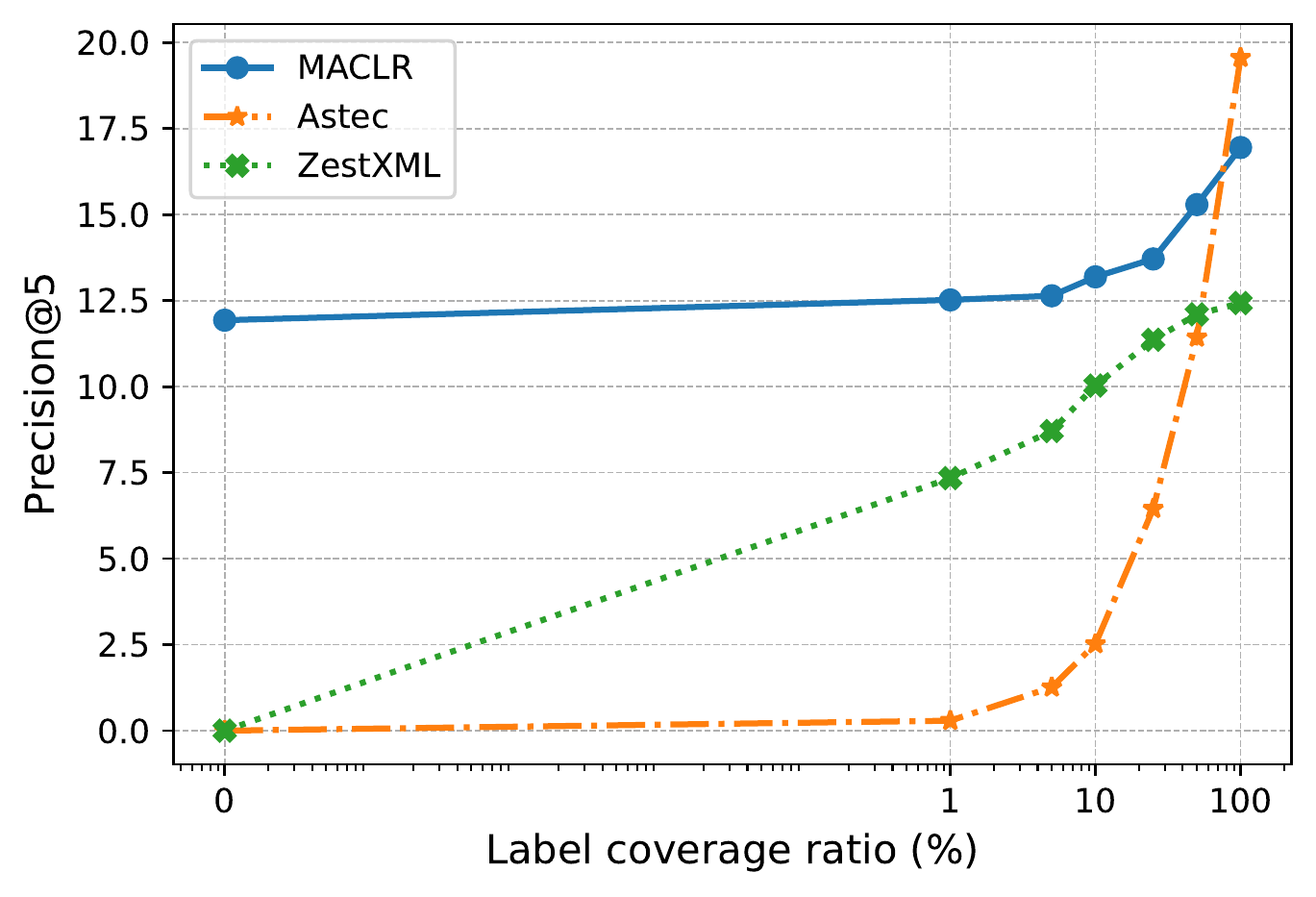}
    \caption{Precision@5.}
    \label{fig:prec}
    \end{subfigure}
    \hspace{2em}
  \begin{subfigure}[ht]{0.4\textwidth}
    \centering
    \includegraphics[width=\textwidth,
    trim={0in 0in 0in 0in},
    clip=false,bb=0 0 384 268]{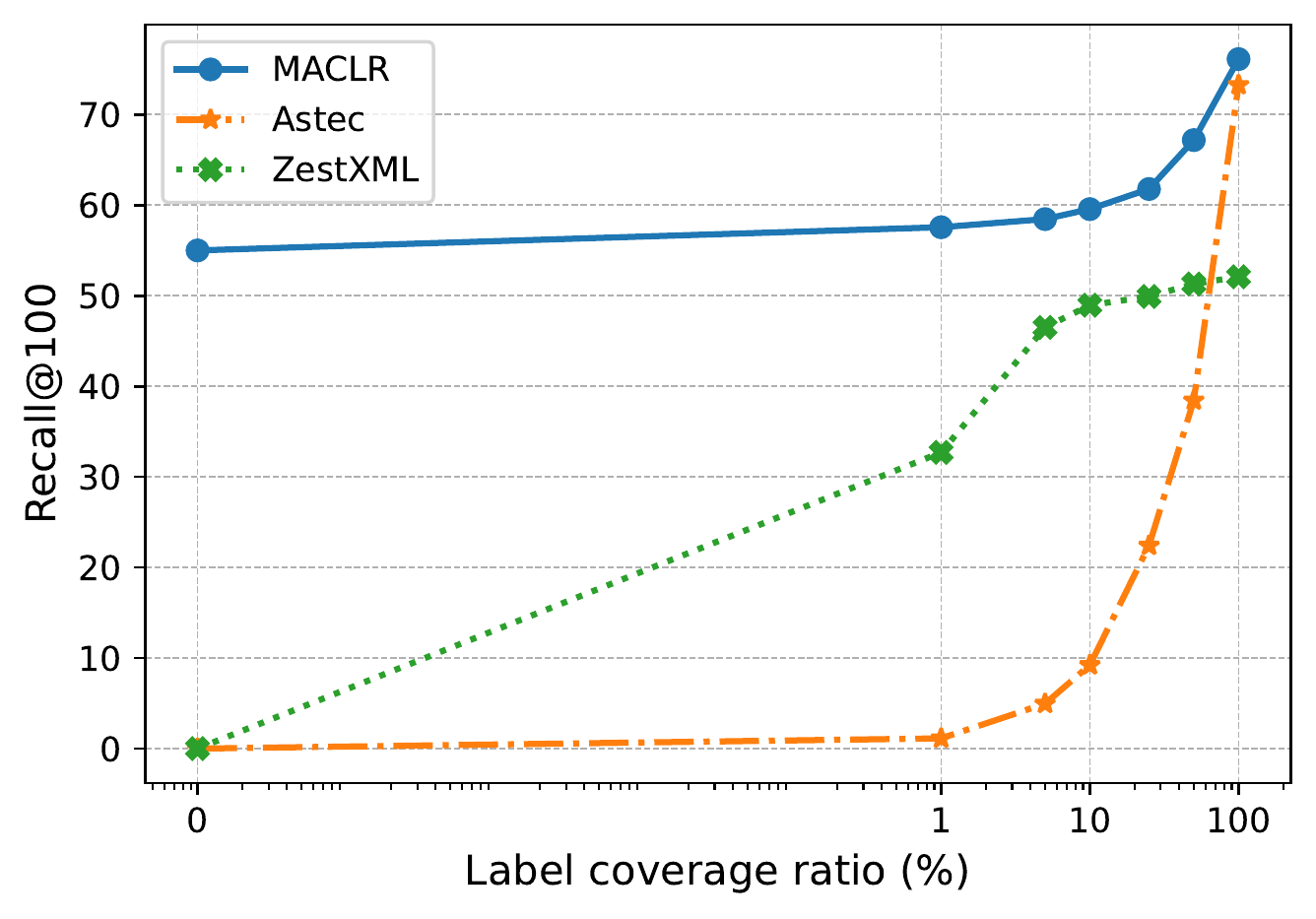}
    \caption{Recall@100.}
    \label{fig:recall}
    \end{subfigure}
  \caption{
    Performance of three representative \xmc methods on the LF-Amazon-131K at different ratios of label coverage.
    A subset of (instance, label) pairs covering $[0, 1, 5, 10, 25, 50, 100](\%)$ of the whole label set is sampled.
  }
  \label{fig:preci}
\end{figure}

However, there are several limitations of this \xmc setting.
First of all, due to the assumption of full label coverage, it is typical in \xmc approaches to simply treat labels as IDs for classification
and thus they are restricted to making predictions within observed labels.
This assumption is unrealistic since the label set usually keeps growing over time, e.g., newly added websites or products which are absent during training yet crucial for applications such as recommendation and advertising.
Besides, collecting labeled pairs is time-consuming, expensive and sometimes infeasible, for example, 
launching an e-commerce system in the emerging locale, where no user behavioral signals are avaiable.
In spite of these constraints, most existing methods~\cite{dahiya2021deepxml,you2019attentionxml,mittal2021eclare, dahiya2021siamesexml} followed this \xmc setup. 
It can be seen in Figure \ref{fig:preci} that Astec~\cite{dahiya2021deepxml}, one of the state-of-the-art extreme classifiers, is incapable of handling the scenario without supervision, which leads to zero performance in both Precision@5 and Recall@100. 
Moreover, the increasing trend in Astec's performance along with the label ratio suggests that it depends highly on the supervision level and is hard to generalize to unseen labels.
This motivates us to investigate how to design an effective XMC model with zero supervision. 

In this paper, we consider an essential yet under-explored \xmc setting, called Extreme Zero-shot XMC (\setting).
As depicted in Figure \ref{fig:scenarios}, we can access raw text of both instances and labels but do not know their corresponding relations in \setting.
Moreover, we do not make any assumption on the label coverage, so the labels in the testing set may or may not appear in the training stage.
An extension to \setting with a limited number of training pairs, Few-shot XMC~(\fsxmc), is also taken into account in our paper.
Either \setting or \fsxmc occurs frequently in the real world since informative and abundant (instance, label) pairs are never easy to obtain.
Also, it is more practical and worthwhile to reduce labor for manual anotation by solving problems under \setting.
For example, the unsupervised model can help narrow the tagging space remarkably by selecting a small number of candidate labels for products in the e-commerce domain for efficient annotation.
 Note that generalized zero-shot XMC~(\gzxmc) proposed in a recent work \cite{gupta2021generalized} can be regarded as a special case of \setting. \gzxmc allows that the set of test labels is not completely overlapped with training labels but still requires supervision from positive pairs, as shown in Figure~\ref{fig:scenarios}.
 From Figure~\ref{fig:preci}, we can observe that \zestxml~\cite{gupta2021generalized} designed for \gzxmc also suffers the issue of no supervision although it can improve the precision and recall significantly under a low label coverage ratio compared with other standard XMC methods such as Astec~\cite{dahiya2021deepxml}.


A natural question then arises: how should we deal with \setting problems?
Despite the name, \setting is barely easy to tackle. 
Fortunately, although dedicated supervision signals are lacking,
raw text of instances and labels, e.g., product descriptions and categories, are still accessible in \setting.
Thus it is of vital importance to effectively leverage self-information of these data to train a model for classification.
To overcome challenges encountered in \setting, we turn to solving the problem from a different perspective without learning classifiers explicitly.
In particular, XMC can be cast into a problem which learns a sentence encoder $\gE$ to map instances and labels into dense embeddings, and predictions are made through approximate nearest neighbor
search algorithms in the latent space~\cite{shrivastava2014asymmetric}.
Motivated by recent progresses in self-supervised learning~\cite{gao2021simcse,chen2020simple,he2020momentum,devlin2018bert}, we propose \textbf{\ours} (\textbf{M}ulti-scale \textbf{A}daptive \textbf{C}lustering \& \textbf{L}abel \textbf{R}egularization), a two-stage pre-training procedure with those unpaired raw data to obtain a sentence encoder $\gE$ under \setting.
Specifically, the Inverse Cloze Task~(ICT)~\cite{lee2019latent} is adopted to construct (context, title) pairs as the unsupervised anlog of (instance, label) pairs.
To make full use of data's self-information, we leverage a multi-scale approach, which conducts clustering for instances with adaptive sizes along with the training process and treats those instances belonging to the same cluster as positive examples in contrastive learning.
It allows the encoder to learn from the coarse to the fine scale gradually.
Label information is utilized as a regularization term to separate an instance from its irrelevant labels.
Moreover, self-training is conducted to further improve the encoder on pseudo positive pairs generated by the initial predictions of our model and TF-IDF.
As to \fsxmc, fine-tuning the encoder on a few paired data is sufficient for the performance boost.
Figure \ref{fig:preci} demonstrates that \ours achieves superior performance when no supervision is available and achieves much higher recall than Astec and ZestXML by a large margin, even under the higher label coverage ratio.

Our main contributions are summarized as follows:
\begin{itemize}[topsep=1pt,parsep=1pt,partopsep=1pt]
  \item We propose an essential Extreme Zero-Shot XMC~(\setting) setting without any assumptions on supervision and label coverage, which has not been explored in previous work and is more practical in real applications.
  \item We leverage unlabeled data to pretrain the sentence encoder $\mathcal{E}$ with improved Inverse Cloze Task in Stage I of \ours.
  In particular, multi-scale adaptive clustering and label regularization are proposed to utilize raw text thoroughly. In Stage II, we further self-train the encoder with pseudo positive pairs constructed from $\mathcal{E}$ in Stage I as well as TF-IDF model with complementary information.
  \item Comprehensive experiments are conducted on four public benchmark \setting datasets.
  Results demonstrate that our pre-trained encoder can outperform existing unsupervised baseline methods notably. As an example, \ours achieves Recall@100 of 54.99\%, nearly the same level as Astec~(one of the SOTA XMC methods)~\cite{dahiya2021deepxml}~ trained with a supervised subset covering around $70\%$ labels on LF-Amazon-131K.
  \item \ours can also achieve comparable or even better performance under the few-shot setting than those models heavily dependent on supervised information. For example, \ours is better than the SOTA ZestXML~\cite{gupta2021generalized} in Recall@100 over $20\%$ ($57.55\%$ v.s. $32.69\%$) when fine-tuned on the subset covering $1\%$ labels of LF-Amazon-131K.
\end{itemize}

\section{Related Work}
\paragraph{\bf Extreme multi-label classification}
Various extreme classifiers have been proposed to address the large output space challenge of \xmc problems.
We can broadly categorize them into two groups: 
partitioned-based models with linear classifiers~\citep{prabhu2018parabel,prabhu2014fastxml,yu2020pecos}
that partition labels with hierarchical trees, leading to sub-linear inference time complexity,
and embedding-based methods~\cite{bhatia2015sparse,jain2019slice,guo2019breaking} that learn a classifier for each label and leverage approximated nearest neighbor \citep{malkov2018efficient,guo2016quantization} to index labels in the large output space.
There are also deep learning models such as
	\attnxml~\citep{you2019attentionxml},
	\astec~\citep{dahiya2021deepxml},
	\siamesexml~\citep{dahiya2021siamesexml}, and
	\xrtransformer~\citep{zhang2021fast} that
further improve the accuracy of those linear counterparts with various advanced encoder architectures.
Nevertheless, none of those \xmc methods can handle the \setting setup:
	they not only suffer from the lack of supervised signals,
	but also fail to generalize to unseen cold-start labels in the test set.

The only exception is \zestxml~\citep{gupta2021generalized}, a recently proposed \xmc method
that was designed to address the generalized zero-shot XMC (\gzxmc) problem where a number of labels for prediction are absent during training.
While \zestxml partially resolves the generalization challenge of cold-start labels,
	  just like those conventional \xmc models,
	  \zestxml still depends heavily on a large number of training data with positive (instance, label) pairs.
None of existing methods consider the scenario where no supervision information is available.
\ours, on the other hand, can make satisfactory predictions even \textit{without any paired data} 
and outperform all baseline methods when there is a limited amount of supervision. 

\paragraph{\bf Dense sentence embeddings}
With increasinglg powerful computing resources, training dense sentence embeddings has become popular to learn the semantics directly from raw text~\cite{reimers2019sentence,xiong2020approximate,lu2020twinbert}.
Based on dense embeddings, approaches such as Sentence-BERT~\cite{reimers2019sentence} and TwinBERT~\cite{lu2020twinbert} project an instance and its relevant labels together in a shared latent space.
Since the embeddings of all candidate labels can be precomputed and indexed, the label prediction can be done efficiently with approximate nearest neighbor
search algorithms in the embedding space \cite{shrivastava2014asymmetric,guo2016quantization}.
Such methods can also be extended to applications with millions of labels, 
and can be utilized to produce meaningful embeddings for even unseen sentences due to the generalization of neural networks.

\paragraph{\bf Self-supervised learning techniques}
The past few years have witnessed great promise in self-supervised learning
~\cite{lan2019albert,chen2020simple,he2020momentum,devlin2018bert,khosla2020supervised,gao2021simcse}, 
where a pre-training task is defined using only data's self-information.
Learned representations from the pre-training task can be then leveraged in a wide range of downstream tasks in various domains, 
such as image classification~\citep{chen2020simple,he2020momentum} and object detection~\citep{li2020improving} in computer vision,
and open-domain question answering~\citep{lee2019latent,guu2020realm} in natural language processing.
Specifically, we focus on contrastive approaches for Sentence-BERT~\citep{reimers2019sentence} models in this paper, 
where the intuition is to pull semantically close neighbors
together and push apart non-neighbors via noise contrastive estimation or N-pair losses. 
Various effective pre-training tasks such as Inverse Cloze Task (ICT)~\cite{lee2019latent}
and \simcse~\citep{gao2021simcse} have been shown to improve the performance of Sentence-BERT models.

\section{Problem Formulation}
In this section, we present the problem formulation of \setting.
As introduced before, with $\gX$ and $\gY$ denoting the set of instances and labels respectively,
the general XMC problem can be viewed as learning a scoring function $f: \gX \times \gY \rightarrow \RR$.
$f(\cdot,\cdot)$ maps an (instance, label) pair $(x, y)$ to a similarity score, which is used to make a prediction through approximate nearest neighbor search algorithms.
In previous settings such as \xmc and \gzxmc, a considerable amount of relevant (instance, label) pairs $\{(x_i, y_i)\}$ are available.
On the contrary, in \setting, we have no knowledge about corresponding relations between instances and labels,
but only their raw text, as shown in Figure~\ref{fig:scenarios}.
 In this case, existing approaches that depend on the relevant pairs fail to learn an effective scoring function,
 even with a few paired data under \fsxmc.

Recent progresses in self-supervised learning have shown that
a generalized sentence encoder can be learned through elaborately designed pre-training tasks even without any supervision~\cite{lee2019latent,chang2020pre},
and then adapted to different downstream tasks directly or via slight finetuning.
On the other hand, the scoring function $f$ can be modeled as
\begin{equation}
	\label{eq:score}
	f(x,y)= \langle \gE(x), \gE(y) \rangle,
\end{equation}
where $\gE$ is a sentence encoder producing semantical dense embeddings,
and $\langle \cdot , \cdot \rangle$ is the similarity measurement such as inner product and cosine similarity.
Without loss of generality, inner product is adopted in the paper as the similarity metric between embeddings of instances and labels.
Thus, we formulate the problem as training such an encoder $\gE$ with raw text of $\gX$ and $\gY$ through a pre-training task for \setting.
As to the few-shot scenario \fsxmc where a limited number of pairs are available, we can fine-tune $\gE$ for improvement.
We will introduce the proposed algorithm for training $\gE$ in Section~\ref{sec:method}. 


\section{Method}
\label{sec:method}
In this section, we introduce a two-stage pre-training procedure, \ours, to thoroughly leverage unpaired data with raw text for \setting.
Specifically, we present the general framework in Section \ref{sec:framework}, and then dive into details of two stages, pre-training with the improved Inverse Cloze Task and self-training with pseudo positive pairs,
in Sections \ref{sec:stage-1} and \ref{sec:stage-2} respectively.
\subsection{Framework}
\label{sec:framework}
\begin{figure}[ht]
	\centering
	\includegraphics[width=0.7\textwidth,bb=0 0 1615 1179]{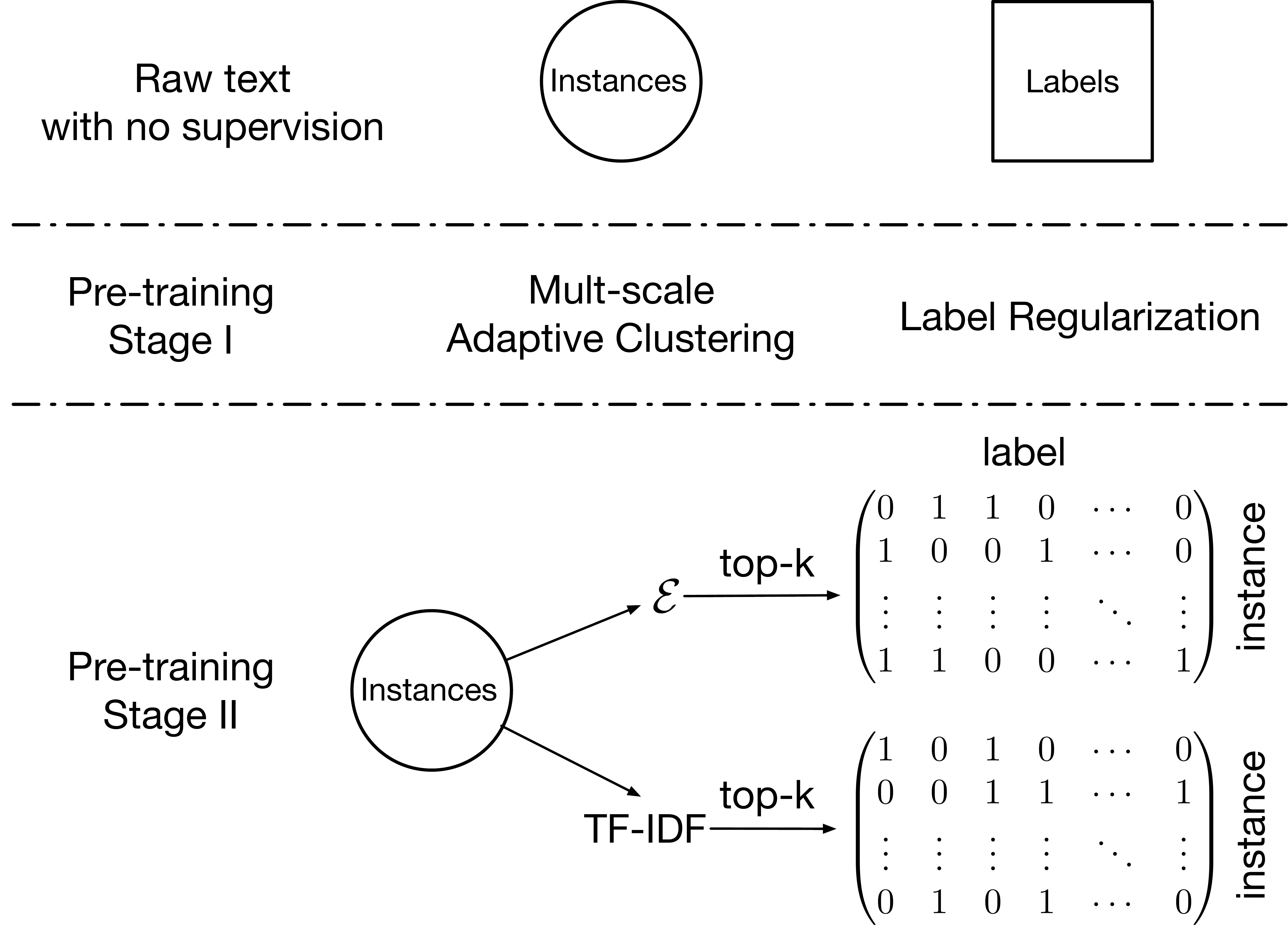}
	\caption{Framework of our pre-training procedure.}
	\label{fig:framework}
  \end{figure}

The framework of our pre-training procedure is shown in Figure \ref{fig:framework}.
\ours  consists of two stages:
\begin{itemize}
	\item Stage I: title-context pairs are constructed for the Inverse Cloze Task, and the encoder $\gE$ is then trained on these pairs together with two proposed techniques, multi-scale adaptive clustering and label regularization.
	\item Stage II: More pseudo positive pairs are crafted using different score functions modeled by the encoder from Stage I and TF-IDF respectively. $\gE$ is further trained on additional pairs
	to improve the encoding performance.
\end{itemize}
We will discuss the details of each component in our pre-training framework in the following sections.

\subsection{Stage I: Pre-training with the improved ICT}
\label{sec:stage-1}
Inverse Cloze Task~\cite{lee2019latent} is a frequently used pre-training task for the sentence encoder.
Specifically, for an instance $x=\{s_1, \dots, s_n\}$ consisting of $n$ sentences,
ICT randomly samples a sentence to serve as the pseudo positive label $\hat{y} = s_i$ where $i \sim [1,n]$.
Then the rest of $x$ is the pseudo instance $\hat{x}=\{s_1,\dots,s_{i-1}, s_{i+1},\dots, s_n\}$.
In XMC, due to the property that the label usually summarizes the instance with one short sentence, which works similarly as the title $s_1$,
we directly utilize (context, title) pairs in the form of ($\hat{x}=\{s_2,\dots,s_n\}, \hat{y}=s_1$).
This contruction is better to work as the analog of the ground truth (instance, label) pairs and capture the semantics of a sentence.
With these pseudo pairs, the contrastive training objective for a mini-batch of $N$ pairs is as follows:
\begin{equation}
	\label{eq:contrastive}
    \mathcal{L}_\text{contrastive} = -\sum_{i=1}^{N}\log \frac{\exp(\gE(\hat{x}_i)\cdot \gE(\hat{y}_i))}{\sum_{j=1}^{N}\exp(\gE(\hat{x}_i)\cdot \gE(\hat{y}_j))}
\end{equation}
Based on ICT, we also develop two techniques, multi-scale adaptive clustering and label reguarization,
to fully leverage the information of unpaired instances and labels.
\subsubsection{Multi-scale Adaptive Clustering}
In the original ICT scheme, we can construct only one positive pair for a particular instance.
It is relatively hard in contrastive learning without enough positive examples,
especially for extreme multi-label classification where
one instance might be associated with more than one label,
and a label is also likely to point to several different instances at the same time.
Thus a question arises naturally: is it possible to construct more positive pairs from purely unpaired raw data
to intergrate richer information into the pre-training process?
We solve it by the unsupervised K-means clustering.
In detail, we divide pseudo (context, title) pairs from ICT into $K$ clusters through K-means based on the embeddings of all instances.
Then if $C(\hat{x}_i)=C(\hat{x}_j)$, i.e., $\hat{x}_i$ and $\hat{x}_j$ belong to the same cluster,  $(\hat{x}_i, \hat{y}_j)$ and $(\hat{x}_j, \hat{y}_i)$ are regarded as positive pairs besides original ICT pairs.
Furthermore, supervised contrastive loss is adopted for training the encoder with a mini-batch of $N$ pairs based on the cluster assignment:
\begin{equation}
	\label{eq:cluster}
    \mathcal{L}_\text{cluster} = \sum_{i=1}^{N}\frac{-1}{|P_{\gY}(i)|}
	\sum_{p \in P_{\gY}(i)} \log \frac{\exp(\gE(\hat{x}_i)\cdot \gE(\hat{y}_p))}{\sum_{j=1}^{N}\exp(\gE(\hat{x}_i)\cdot \gE(\hat{y}_j))}
\end{equation}
Here, $P_{\gY}(i)=\{p\in\{1,\dots, N\}:C(\hat{x}_i)=C(\hat{x}_p)\}$ is the set of indices of all positives for $\hat{x}_i$ in the batch, and $|P_{\gY}(i)|$ is its cardinality.
Minimizing \eqref{eq:cluster} pulls close the representations of instances and their positive labels within the same cluster and pushes away the representations of those from different clusters.

\begin{figure}[ht]
	\centering
	\includegraphics[width=0.6\textwidth,bb=0 0 719 271]{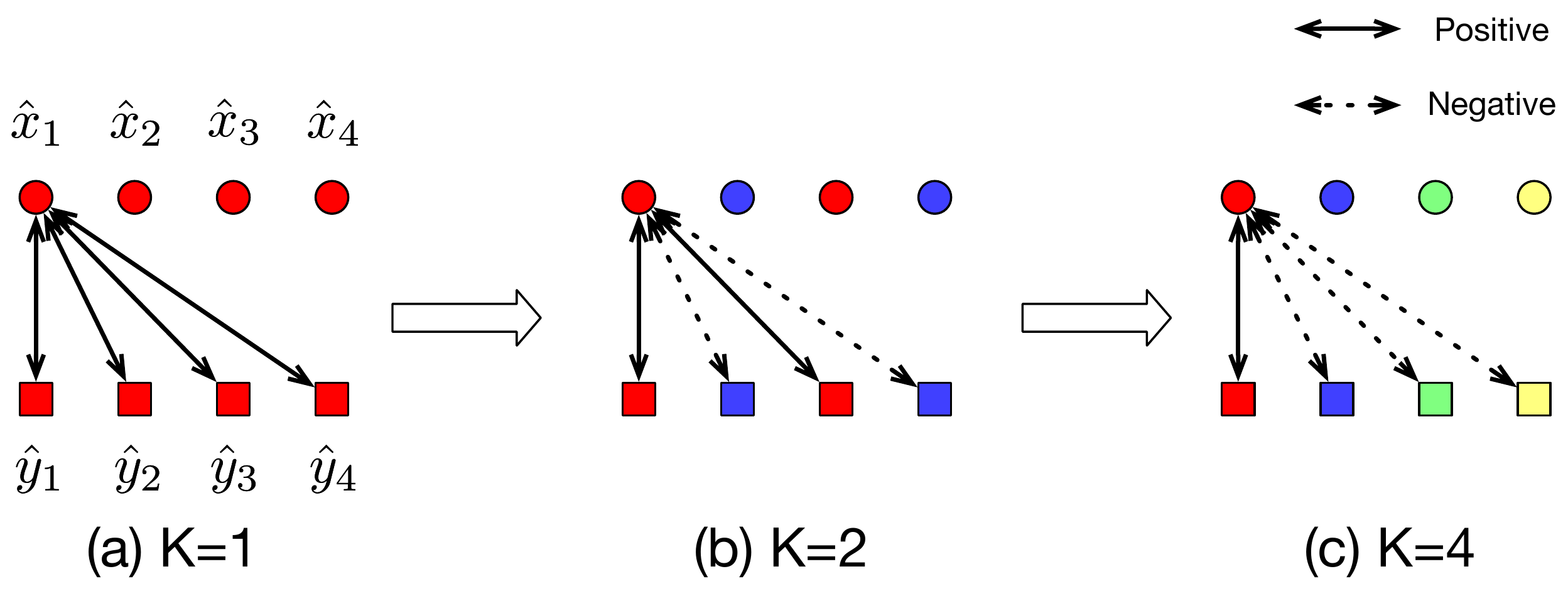}
	\caption{An example of multi-scale adaptive clustering. Here different colors represent different clusters.
	(a) In the beginning, there is only one cluster and $\{\hat{y}_j\}_{j=1}^4$ are all positive labels for $\hat{x}_1$.
	(b) $K$ is doubled to 2 and now $\hat{y}_1$ and $\hat{y}_3$ are positive to $\hat{x}_1$.
	(c) Finally, $K$ is equal to 4 where each instance itself is a cluster, and hence $\hat{x}_1$ only has one positive label $\hat{y}_1$.
	The process is similar for the rest of the instances.}
	\label{fig:ac}
  \end{figure}

Besides, since the ultimate goal is the minimization of \eqref{eq:contrastive},
we propose a multi-scale approach with adaptive training,
which guides the encoder to learn the easier tasks with sufficient positive examples, and then master harder tasks gradually.
This approach allows the encoder to learn from the coarse scale to the fine scale of clustering assignment, and is similar to the idea of curriculum learning~\cite{bengio2009curriculum} to first focus on learning from a subset of simple examples, and expanding to include the remaining harder samples.
Our adaptive training process can be conducted by modifying the cluster size to adjust the task difficulty accordingly.
To be specific, we initialize the cluster assignment with the number of clusters $K_0$, and double the cluster size every $T$ steps.
The cluster assignment is also updated every $T_\text{update}$ steps along with the training of $\gE$.
Such a process lasts for half of the total training steps $T_\text{total}$ to take advantage of positive examples from constructed clusters.
The obtained intermediate encoder from this adaptive procedure is expected to satisfactorily capture the semantics of a sentence
and is ready to deal with the optimization with \eqref{eq:contrastive}.
Then for the rest half of training steps, we turn to the hardest setting treating each instance as one independent cluster,
which exactly falls into the contrastive training objective in \eqref{eq:contrastive}.
Our multi-scale adaptive clustering is illustrated in Figure~\ref{fig:ac}.

\subsubsection{Label Regularization}
\begin{figure}[ht]
	\centering
	\includegraphics[width=0.6\textwidth,bb=0 0 719 271]{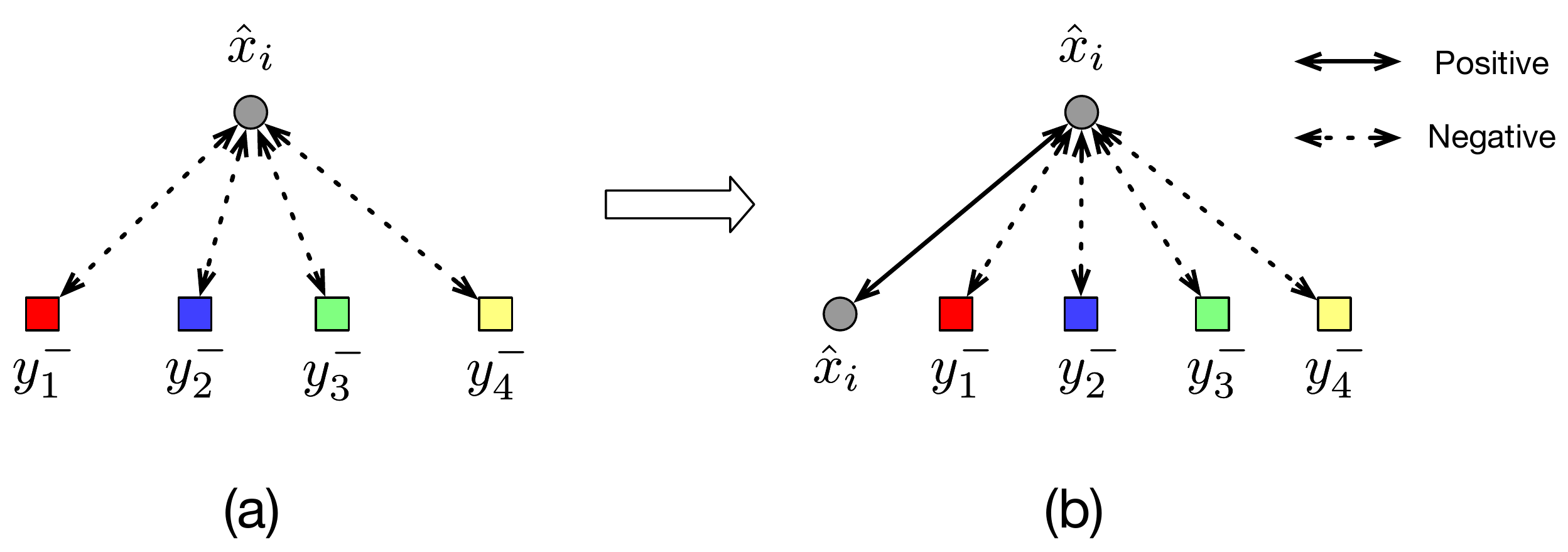}
	\caption{An illustration of label reguarization.
	(a) shows that $\hat{x}_i$ is expected to be far apart from sampled irrelevant labels $\{y_j^-\}_{j=1}^4$,
	while (b) indicates the identical $\hat{x}_i$ is added as a positive example for label regularization.}
	\vspace{-1em}
	\label{fig:lr}
\end{figure}

In addition to leveraging information from the instance side,
we also have access to the raw texts of the whole label set and can utilize them to boost the encoder's performance from the label side.
Intuitively, for a randomly sampled label, with a high probability it is an negative example to the instance of interest.
We can take advantage of this intuition to make the embedding of the instance far apart from its irrelevant labels.
Instead of increasing the distance directly, it is more stable and effective to adopt contrastive losses.
To avoid overfitting, we choose a new positive example for each instance instead of its corresponding pseudo label from ICT which has been used in $\mathcal{L}_\text{cluster}$.
More concretely, $\hat{x}_{i}^+$ is selected exactly the same as $\hat{x}_i$,
since the dropout layer is placed in the standard training of Transformer-based models and can be viewed as a minimal form of data augmentation~\cite{gao2021simcse}.
By feeding the same sentence to the encoder $\gE$, two embeddings with different dropout masks are obtained,
i.e., $\hat{h}_i = \gE(\hat{x}_i, z_i)$ and $\hat{h}_i^+ = \gE(\hat{x}_i^+, z_i^+)$ where $z$ represents a random mask for dropout.
Note that $\hat{h}_i \neq \hat{h}_i^+$ due to the dropout noise,
but they hold similar semantics from the same sentence and thus can be used as a positive pair for contrastive learning.
The procedure of label regularization is depicted in Figure \ref{fig:lr}. At each training step, we sample $M$ real labels from the label set $\gY$,
and the reguarization term is computed as follows:
\begin{equation}
    \mathcal{L}_\text{label} = \sum_{i=1}^{N}-\log \frac{\exp(\hat{h}_i\cdot \hat{h}_i^+)}{\exp(\hat{h}_i\cdot \hat{h}_{i}^+)+\sum_{j=1}^{M}\exp(\hat{h}_i\cdot \gE(y_j^-))}
\end{equation}
Through minimizing $\gL_\text{label}$, the encoder learns to pull the instance away from its irrelevant labels and incorporate the dropout augmentation at the same time.
Together with $\gL_\text{cluster}$, we have the final objective function for pre-training in the Stage I as
\begin{equation}
	\gL = \gL_\text{cluster} + \gL_\text{label}.
\end{equation}
We present the detailed implementation in Algorithm \ref{alg:stage-i}.

  \begin{algorithm}[ht]
	\caption{Pre-training procedure of \ours}
	\label{alg:stage-i}
	\begin{algorithmic}[1]
		\Require Raw text of instances and labels $(\gX, \gY)$, the sentence encoder $\gE$, batch size $N$ and $M$, training step parameters $T_K$, $T_\text{update}$ and $T_\text{total}$, initial cluster size $K_0$, \# of top candidates $k$
		\Ensure A pre-trained sentence encoder $\gE$
	\LeftComment{Stage I: Pre-training with the improved ICT}
	\State Construct ICT (context, title) pairs from raw texts in $\gX$
	\State Feed the context for each pair into the encoder $\gE$ and cluster them into $K=K_0$ clusters via k-means
	\For {$t=1,\dots,T_\text{total}$}
		\State Sample a mini-batch of pseudo pairs of size $N$ and a mini-batch of real labels of size $M$
		\State Compute the loss: $\mathcal{L}= \gL_\text{cluster} + \gL_\text{label}$
		\State Train the encoder by minimizing $\mathcal{L}$
		\If {$t$ mod $T_K = 0$ and $t < T_\text{total}/2$ }
		\State $K = K*2$
		\EndIf

		\If {$t$ mod $T_\text{update} = 0$ and $t < T_\text{total}/2$}
		\State Feed raw texts of $\gX$ again into $\gE$, and update current cluster assignment via k-means with the cluster number $K$
		\EndIf

		\If {$t \geq T_\text{total}/2$}
		\State Treat each instance as an independent cluster
		\EndIf
	\EndFor
	\LeftComment{Stage II: Self-training with multi-viewed pseudo pairs}
	\State Construct pseudo pairs $(\gX_\text{pseu}, \gY_\text{pseu})$ by selecting top-$k$ candidate labels with the similarity metric in \eqref{eq:score} on the encoder $\gE$ and TF-IDF respectively
	\State Train the encoder $\gE$ for $T_\text{total}$ steps by minimizing \eqref{eq:cluster}
	\end{algorithmic}
	\end{algorithm}
\subsection{Stage II: Self-training with multi-viewed pseudo pairs}
\label{sec:stage-2}
After the pre-training procedure in Section~\ref{sec:stage-1}, we can obtain an intermediate encoder $\gE_{I}$.
But are there any ways to further improve the encoder?
On the one hand, inspired by self-training in semi-supervised learning~\cite{yalniz2019billion,xie2020self,he2019revisiting,zoph2020rethinking},
$\gE_{I}$ can be leveraged to make predictions on those unpaired training instances themselves,
to generate pseudo positive pairs.
These pseudo pairs are much better than random guessing and can serve as a distinct view from ICT pairs.
On the other hand, similar pseudo pairs can be constructed by other unsupervised methods such as TF-IDF, which also provide different and complementary information about the instance.

With multi-viewed pseudo positive pairs,
we can conduct further training on the encoder in State II from a new perspective and self-improve $\gE_{I}$.
The detailed process works as follows:
\begin{itemize}[noitemsep,topsep=1pt,parsep=1pt,partopsep=1pt]
	\item[1)] Compute the similarity score in \eqref{eq:score} using $\gE_{I}$ for each training instance i,
	and select labels with top-k maximum scores as its pseudo labels;
	\item[2)] Generate labels similarly with TF-IDF, except that $\gE(x)$ and $\gE(y)$ are replaced with their TF-IDF vectors;
	\item[3)] Mix pseudo positive pairs from 1) and 2) together, and train $\gE_{I}$ on them with \eqref{eq:cluster}.
	Here the positive index set for the instance $\hat{x}_i$ is defined as $P_{\gY}(i) = \{p | \hat{x}_p = \hat{x}_i\}$.
\end{itemize}

The whole pre-training procedure of \ours is shown in Algorithm \ref{alg:stage-i}.
For \fsxmc, we simply fine-tune the encoder $\gE$ from \ours on available positive pairs for several steps by minimizing \eqref{eq:contrastive}.

\section{Experimental Results}
\subsection{Experimental Settings}

\paragraph{\bf Datasets}
We evaluate \ours on $4$ public \xmc benchmark datasets~\citep{Bhatia16,gupta2021generalized}
where raw text of instances and labels are available.
These datasets are derived from real-world applications,
	ranging from item-to-item recommendation (\amznsmall, \amznlarge),
	to Wikipedia articles category/title tagging (\wikismall, \wikilarge).
All datasets are available in the
	Extreme Classification Repository~\citep{Bhatia16}\footnote{\url{http://manikvarma.org/downloads/XC/XMLRepository.html}},
except for \amznlarge, which is available from the
	\zestxml paper~\cite{gupta2021generalized}\footnote{\url{https://github.com/nilesh2797/zestxml}}.
Detailed dataset statistics are presented in Table \ref{tab:statistics}.

\begin{table}[!ht]
  \centering
  \caption{
	  Dataset statistics.
	  $\mathbf{N_\text{train}}$, $\mathbf{N_\text{test}}$ and $\mathbf{N_\text{label}}$
	  are the number of training points, test points, and labels respectively.
    $\mathbf{D_\text{BoW}}$ is the dimensionality of Bag-of-Words~(BoW) features.
  }
  \label{tab:statistics}
  \begin{tabular}{l|rrrr}
  \toprule
  \textbf{Dataset} & $\mathbf{N_\text{train}}$ & $\mathbf{N_\text{test}}$ & $\mathbf{N_\text{label}}$ & $\mathbf{D_\text{BoW}}$ \\
  \midrule
  \amznsmall &   294,805 &   134,835 & 131,073 & 80,000 \\
  \wikismall &   693,082 &   177,515 & 312,330 & 80,000\\
  \wikilarge & 1,813,391 &   783,743 & 501,070 & 500,000\\
  \amznlarge &   914,179 & 1,465,767 & 960,106 &  1,000,000\\
  \bottomrule
  \end{tabular}
\end{table}
\paragraph{\bf Evaluation Protocol}
We consider two evaluation setups: Extreme Zero-shot Learning of \xmc (\setting) and Few-shot Learning of \xmc (\fsxmc).
\setting is a fully unsupervised learning setup where no positive (instance, label) pairs are available.
The only available information is the raw text of training instances and the whole label set.
\fsxmc is a semi-supervised learning setup where only very few positive (instance, label) pairs in the training set are available.
Regardless of the learning procedure, all models are evaluated on the same test set for fair comparison.

We evaluate the models' performance with precision@k (P@k, $k\in\{1,3,5\}$) and recall@k (R@k, $k\in\{1,3,5,10,100\}$),
which are two commonly-used evaluation metrics in the \xmc literature~\citep{reddi2019stochastic,chang2021extreme}.
In detail, $P@k$ and $R@k$ are defined as follows:
\begin{equation}
  P@k = \frac{1}{k}\sum_{i\in\text{rank}_k(y')} y_i,  \quad R@k = \frac{1}{\sum_l y_l}\sum_{i\in\text{rank}_k(y')} y_i,
\end{equation}
where $y\in \{0, 1\}^L$ and $y' \in R^L$ are the ground true vector and the prediction vector respectively.
Here $\text{rank}_k$ returns the indices of the top-$k$ highest elements.

\paragraph{\bf Baseline Methods}
For \setting, we compare our method with the following unsupervised learning algorithms:
\begin{itemize}[noitemsep,topsep=1pt,parsep=1pt,partopsep=1pt,leftmargin=*]
	\item \tfidf~\cite{rajaraman2011mining}, which represents instances and labels by sparse TF-IDF features and retrieves top labels for each instance based on the similarity of TF-IDF features;
	\item \xrl~\cite{yu2020pecos}, a hierarchical linear model trained with pseudo positive pairs constructed from \tfidf;
	\item \glove~\citep{pennington2014glove}, which adopts dense average word embeddings with the dimension of 300 trained on co-occurrence statistics to measure similarity between instances and labels;
	\item Sentence-BERT (\sbert)~\citep{devlin2018bert,reimers2019sentence}, a sentence encoder modeled as a Siamese-Transformer to derive semantically meaningful embeddings for instances and labels;
  \item Paraphrase MPNet (\mpnet)~\cite{song2020mpnet}, another Sentence-BERT model designed for searching sentence paraphrases;
	\item SimCSE~\cite{gao2021simcse}, a Transformer pre-trained with the contrastive objective using dropout noise as augmentation;
	\item ICT~\cite{lee2019latent}, another Siamese-Transformer pre-trained with the contrastive objective using (context, title) pairs.
\end{itemize}
Note that \sbert and \mpnet are pre-trained on external multi-task learning datasets with extra supervision.
In contrast, \simcse and \ict are fully unsupervised pre-rained Siamese-Transformers on the specific \xmc dataset only.

For the setting of \fsxmc, as few-shot (instance, label) pairs are available,
we additionally compare fine-tuned \ours with competitive \xmc approaches,
including \astec~\cite{dahiya2021deepxml}, \siamesexml~\cite{dahiya2021siamesexml},
and \zestxml~\citep{gupta2021generalized}.
It should be noticed that \zestxml is the leading \xmc method that improves performance on few-shot labels.
We also take into account of \sbert~\cite{reimers2019sentence} with further fine-tuning to demonstrate the effectiveness of our pre-training procedure.

\paragraph{\bf Hyper-parameters}
We use a Siamese Transformer model to embed both instances and labels.
The encoder consists of a 12 layers BERT-base model, topped
with a linear head projecting hidden state of the [CLS] token into a $512$-dimensional embedding.
The sequence length of the instance and the label is set to be 288 and 64 respectively.
We pre-train the model on eight V100 GPUs for 100,000 steps with an Adam optimizer
and batch size of 32 per GPU in both Stage I and Stage II.
This pre-training process takes about 1 day.
We adopt an initial learning rate $1\times 10^{-5}$ with the warm-up ratio 0.1, followed by a linear learning rate decay.
For fine-tuning, the learning rate of Adam is set to $5\times 10^{-6}$
with 2000 training steps for the $1\%$ label ratio and 10K training steps for the $5\%$ label ratio.
In the Stage I, we use the initial cluster size $K=2048$ and set $T_K=10000$ and $T_\text{update}=5000$.
As to the Stage II, top 3 ranked labels from predictions of the encoder and \tfidf are selected to
constitute the pseudo set for self-training.

For hyper-parameters of all baselines, we follow their default setups.
All experiments are conducted on the AWS p3dn.24xlarge instance,
consisting of 96 Intenl Xeon CPUs with 768 GB of RAM and
8 Nvidia V100 GPUs with 32 GB of memory each.

\subsection{Zero-Shot Learning}
\begin{table}[!ht]
  \caption{Extreme Zero-shot Learning (\setting) comparison of different unsupervised methods.}
  \label{tab:unsup}
  \centering
  \begin{tabular}{l|rrr|rrrrr}
    \toprule
    \multirow{2}{*}{Method}   & \multicolumn{3}{c|}{Precision} &\multicolumn{5}{c}{Recall} \\
    & @1 & @3 & @5 & @1 & @3 & @5 & @10 & @100 \\
    \midrule
    \multicolumn{9}{c}{LF-Amazon-131K} \\
    \midrule
    \tfidf                    & 12.38 & 11.50 &  9.14 &  6.91 & 18.14 & 23.21 & 29.32 & 45.04 \\
    \xrl                      &  7.56 &  7.84 &  7.30 &  4.05 & 12.11 & 18.32 & 29.17 & 40.39 \\
    \glove                    &  3.67 &  2.78 &  2.15 &  2.05 &  4.33 &  5.44 &  7.23 & 14.17 \\
    \sbert                    &  1.86 &  1.44 &  1.14 &  1.01 &  2.22 &  2.88 &  4.01 & 10.18 \\
    \mpnet        			  & 13.94 & 11.41 &  8.82 &  7.82 & 18.08 & 22.58 & 27.91 & 43.39 \\
    \simcse       			  & 10.13 &  8.61 &  6.69 &  5.61 & 13.39 & 16.84 & 21.27 & 35.81 \\
    \ict          			  & 13.82 & 11.41 &  8.90 &  7.76 & 18.09 & 22.80 & 28.94 & 47.40 \\
	\ours (ours)			  & {\bf 18.13} & {\bf 15.42} & {\bf 11.93} & {\bf 10.35} & {\bf 24.45} & {\bf 30.43} & {\bf 37.28} & {\bf 54.99} \\
    \midrule
    \midrule
    \multicolumn{9}{c}{LF-WikiSeeAlso-320K} \\
    \midrule
    \tfidf                    & 10.71 &  8.90 &  7.15 &  5.92 & 13.03 & 16.48 & 21.60 & 42.55 \\
    \xrl                      &  4.73 &  4.27 &  3.90 &  2.23 &  5.83 &  8.64 & 14.18 & 36.93 \\
    \glove                    &  3.86 &  2.76 &  2.21 &  2.12 &  4.11 &  5.22 &  6.95 & 15.33 \\
    \sbert                    &  1.71 &  1.27 &  1.06 &  1.08 &  2.16 &  2.90 &  4.17 & 10.76 \\
    \mpnet                    & 13.75 & 11.93 &  9.58 &  8.14 & 17.77 & 22.21 & 28.11 & 45.91 \\
    \simcse                   &  9.03 &  6.64 &  5.22 &  4.99 &  9.89 & 12.34 & 15.93 & 30.11 \\
    \ict                      & 10.76 & 10.05 &  8.12 &  6.12 & 14.32 & 18.05 & 23.01 & 39.77 \\
	\ours (ours)              & {\bf 16.31} & {\bf 13.53} & {\bf 10.78} &  {\bf9.71} & {\bf20.39} & {\bf25.37} & {\bf32.05} & {\bf53.83} \\
    \midrule
    \midrule
    \multicolumn{9}{c}{LF-Wikipedia-500K} \\
    \midrule
    \tfidf                    & 20.30	& 12.98	&  9.96	&  7.25	& 12.91	& 15.98	& 20.31	& 38.16 \\
    \xrl                      & 10.67	&  8.77	&  7.61	&  3.69	&  8.58	& 12.11	& 19.80	& 31.02 \\
    \glove                    &  2.19	&  1.52	&  1.23	&  0.85	&  1.66	&  2.18	&  3.10	&  8.52 \\
    \sbert                    &  0.17	&  0.15	&  0.13	&  0.05	&  0.13	&  0.18	&  0.30	&  1.29 \\
    \mpnet					  & 22.46	& 12.87	&  9.49	&  8.74	& 14.07	& 16.76	& 20.64	& 34.72 \\
    \simcse                   & 14.32	&  6.84	&  4.55	&  4.24	&  8.03	& 11.26	& 14.35	& 27.68 \\
    \ict                      & 17.74	&  9.67	&  7.06	&  7.35	& 11.60	& 13.84	& 17.19	& 31.08 \\
	\ours (ours)       		  & {\bf28.44}	& \bf 17.75	&  \bf 13.53	&\bf 10.40	&\bf 18.16	&\bf 22.38	&\bf 28.52	&\bf 50.09 \\
    \midrule
    \midrule
    \multicolumn{9}{c}{LF-Amazon-1M} \\
    \midrule
    \tfidf                    &  7.68	&  9.20	&  7.23	&  5.61	& 19.30	& 24.92	& 31.76	& 51.79 \\
    \xrl                      &  5.19	&  5.48	&  5.26	&  3.63	& 11.30	& 17.94	& 31.18	& 43.79 \\
    \glove                    &  4.05	&  4.07	&  3.07	&  2.91	&  8.42	& 10.44	& 12.90	& 21.18 \\
    \sbert                    &  2.82	&  2.87	&  2.13	&  2.03	&  5.91	&  7.21	&  8.80	& 14.22 \\
    \mpnet		              &  8.29   &  8.87	&  6.80	&  6.04	& 18.64	& 23.51	& 29.35	& 46.15 \\
    \simcse                   &  3.33	&  3.69	&  2.74	&  2.38	&  7.66	&  9.38	& 11.43	& 18.54 \\
    \ict                      &  8.66	&  9.26	&  7.13	&  6.30	& 19.45	& 24.60	& 30.73	& 48.42 \\
	\ours (ours)              &  \bf9.58	& \bf10.41	& \bf 8.03	& \bf 7.38	& \bf22.01	& \bf27.72	& \bf34.48	& \bf55.23 \\
    \bottomrule
  \end{tabular}
  \vspace{-1em}
\end{table}

In this section, we focus on extreme zero-shot learning (\setting),
	where no real positive (instance, label) pairs are accessible.
Table \ref{tab:unsup} presents detailed performance of precision and recall on all four datasets.
Our proposed \ours consistently outperforms all comparing baselines with a large margin for all four datasets.
Compared to the leading sparse method \tfidf,
	\ours has an average of 5.3\% and 9.1\% absolute improvement in Precision@1 and Recall@100, respectively.
Compared to the leading neural model \mpnet,
	\ours has an average of 3.5\% and 10.9\% absolute improvement in Precision@1 and Recall@100, respectively.
Speaking of sparse lexical matching approaches, \tfidf remains a tough-to-beat unsupervised baseline.
Specifically, \tfidf performs better than many BERT variants (e.g., \sbert, \simcse, \ict),
which is aligned with the finding in recent zero-shot dense retrieval literature~\citep{thakur2021beir,anonymous2022contrastive}.
This suggests the importance of designing proper self-supervised learning tasks for Transformer models
in unsupervised \setting setup.
Also note that \xrl is based on \tfidf vectors whereas the noise from pseudo pairs makes it even inferior to the original \tfidf method.

As for pre-trained \sbert models, on the other hand, only \mpnet shows comparable performance with \tfidf.
\mpnet remains competitive because it was trained on a large supervised corpus (out-of-domain) to learn
semantics between paraphrasing sentences.
Thus, \mpnet should be viewed as a multi-task learning baseline with extra supervision.
However, \ours is significantly better than \mpnet with an average improvement of 3.5\% in P@1 and over 10\% in R@100.
Furthermore, \ours also outperforms its counterparts which are trained with effective pre-training tasks such as \simcse and \ict on the target dataset,
showing the effectiveness of pre-training strategies like multi-scale adaptive clustering in \ours.
Overall, results in Table \ref{tab:unsup} demonstrates that
\ours is capable to learn informative embeddings and to make useful predictions even with no supervision.
We will investigate each component in \ours in Section \ref{sec:ablation} thoroughly.

\subsection{Few-Shot Learning}
We further conduct few-shot learning (\fsxmc) experiments in which different learning algorithms can access a limited number of positive (instance, label) pairs.
To simulate the scenario of few-shot learning,
we first manually sample a small ratio of labels, then collect all their positive instances from the training set
as the final subset of positive (instance, label) pairs for model training.
Results of \fsxmc methods fine-tuned with 1\% and 5\% labels are shown in Tables \ref{tab:few-shot-1} and \ref{tab:few-shot-5} respectively.

\begin{table}[!ht]
  \caption{Results of few-shot learning \xmc (\fsxmc) where the training subset covers 1\% labels from the whole set.}
  \label{tab:few-shot-1}
  \centering
\begin{tabular}{l|rrr|rrrrr}
    \toprule
\multirow{2}{*}{Method}   & \multicolumn{3}{c|}{Precision} &\multicolumn{5}{c}{Recall} \\
& @1 & @5 & @10 & @1 & @3 & @5 & @10 & @100 \\
    \midrule
    \multicolumn{9}{c}{LF-Amazon-131K} \\
    \midrule
    \xrl	                  &  1.53	&  0.57	&  0.36	&  0.67	&  0.75	&  0.78	&  0.81	&  0.92 \\
    \astec                    &  0.94	&  0.44	&  0.29	&  0.55 &  0.78	&  0.84	&  0.91	&  1.13 \\
   	\siamesexml               &  1.45	&  0.56	&  0.35	&  0.84	&  0.96	&  1.00	&  1.03	&  1.16 \\
    \zestxml                  & 10.10   &  9.19	&  7.34	&  5.63	& 14.46	& 18.61	& 23.73	& 32.69 \\
    \sbert                    & 12.64	&  9.82	&  7.80	&  6.97	& 15.34	& 19.74	& 25.33	& 43.53 \\
    \mpnet		              & 14.78	& 11.55	&  8.97	&  8.28	& 18.24	& 22.84	& 28.54	& 45.89 \\
	\ours (ours)              & \bf18.74	& \bf16.07	& \bf12.52	& \bf10.73	& \bf25.44	& \bf31.89	& \bf39.17	& \bf57.55 \\
    \midrule
    \midrule
    \multicolumn{9}{c}{LF-WikiSeeAlso-320K} \\
    \midrule
    \xrl	                  &  1.24	&  0.57	&  0.37	&  0.42	&  0.58	&  0.63	&  0.68	&  0.76 \\
    \astec                    &  1.25	&  0.60	&  0.41	&  0.69	&  0.98	&  1.11	&  1.27	&  1.56 \\
    \siamesexml               &  1.81	&  0.75	&  0.48	&  1.03	&  1.26	&  1.33	&  1.41	&  1.67 \\
    \zestxml                  &  8.74	&  6.78	&  5.41	&  4.68	&  9.70	& 12.21	& 15.73	& 24.98 \\
    \sbert                    & 16.30	& 12.62	& 10.08	&  9.30	& 18.92	& 23.78	& 30.40	& 52.92 \\
    \mpnet			          & 17.14	& 12.64	&  9.96	&  9.97	& 18.98	& 23.45	& 29.67	& 50.75 \\
	\ours (ours)              & \bf19.09	& \bf14.57	& \bf11.53	& \bf11.39	& \bf22.34	& \bf27.63	& \bf34.81	& \bf57.92 \\
    \midrule
    \midrule
    \multicolumn{9}{c}{LF-Wikipedia-500K} \\
    \midrule
    \xrl	                  &  2.95	&  1.19	&  0.75	&  0.62	&  0.74	&  0.76	&  0.79	&  0.84 \\
    \astec                    &  2.85	&  1.16	&  0.73	&  1.46	&  1.75	&  1.84	&  1.92	&  2.08 \\
    \siamesexml               &  2.72	&  1.15	&  0.73	&  1.39	&  1.73	&  1.84	&  1.93	&  2.09 \\
    \zestxml                  & 23.86	& 14.97	& 11.31	&  7.19	& 13.00	& 16.03	& 20.13	& 29.95 \\
    \sbert	                  & 32.09	& 20.50	& 15.78	& 10.94	& 19.46	& 24.12	& 30.94	& 55.94 \\
    \mpnet		 	          & 34.58	& 22.02	& 16.86	& 11.96	& 21.32	& 26.30	& 33.53	& 57.78 \\
	\ours (ours)              & \bf44.27	& \bf28.46	& \bf21.83	& \bf15.14	& \bf27.04	& \bf33.33	& \bf42.03	& \bf67.95 \\
    \midrule
    \midrule
    \multicolumn{9}{c}{LF-Amazon-1M} \\
    \midrule
    \xrl	                  &  0.51 &  0.20 &  0.12 &  0.36 &  0.42 &  0.43 &  0.45 &  0.49 \\
    \astec                    &  0.49 &  0.59 &  0.12 &  0.34 &  0.40 &  0.42 &  0.44 &  0.49 \\
    \siamesexml               &  0.60 &  0.73 &  0.15 &  0.41 &  0.46 &  0.48 &  0.49 &  0.53 \\
    \zestxml                  &  5.07 &  5.89 &  4.38 &  3.68 & 12.31 & 15.04 & 17.80 & 22.51 \\
    \sbert	                  &  6.56 &  6.93 &  5.68 &  4.35 & 18.29 & 24.72 & 28.69 & 48.52 \\
    \mpnet			          &  8.87 & 10.34 &  7.56 &  6.78 & 20.11 & 26.14 & 31.98 & 50.48 \\
	\ours (ours)              & \bf10.37 & \bf11.23 &  \bf8.58 &  \bf7.57 & \bf23.55 & \bf29.60 & \bf36.71 & \bf56.44 \\
    \bottomrule
\end{tabular}
\end{table}
\begin{table}[!ht]
  \caption{Results of few-shot learning \xmc (\fsxmc) where the training subset covers 5\% labels from the whole set.}
\label{tab:few-shot-5}
  \centering
  \begin{tabular}{l|rrr|rrrrr}
    \toprule
       \multirow{2}{*}{Method}   & \multicolumn{3}{c|}{Precision} &\multicolumn{5}{c}{Recall} \\
      & @1 & @5 & @10 & @1 & @3 & @5 & @10 & @100 \\
    \midrule
    \multicolumn{9}{c}{LF-Amazon-131K} \\
    \midrule
    \xrl	                  &  5.09	&  2.09	&  1.32	&  2.36	&  2.86	&  3.02	&  3.18	&  3.74 \\
    \astec                    &  3.94	&  1.92	&  1.26	&  2.31	&  3.34	&  3.66	&  4.00	&  4.96 \\
    \siamesexml               &  5.36	&  2.23	&  1.41	&  3.15	&  3.89	&  4.08	&  4.27	&  4.82 \\
    \zestxml                  & 12.33	& 10.99	&  8.71	&  6.84	& 17.19	& 21.97	& 28.10	& 46.49 \\
    \sbert	                  & 15.47	& 12.24	&  9.64	&  8.63	& 19.23 & 24.40	& 30.82	& 49.22 \\
    \mpnet			          & 15.03	& 11.88	&  9.28	&  8.47	& 18.74	& 23.69	& 29.93	& 48.84 \\
	\ours (ours)              & \bf19.56	& \bf16.19	& \bf12.64	& \bf11.15	& \bf25.65	& \bf32.18	& \bf39.63	& \bf58.45 \\
    \midrule
    \midrule
    \multicolumn{9}{c}{LF-WikiSeeAlso-320K} \\
    \midrule
    \xrl                 	  &  4.69	&  2.20	&  1.46	&  1.82	&  2.41	&  2.63	&  2.82	&  3.42 \\
    \astec                    &  5.90	&  2.80	&  1.86	&  3.26	&  4.49	&  4.95	&  5.49	&  6.83 \\
    \siamesexml               &  6.83	&  3.15	&  2.06	&  3.88	&  5.15	&  5.56	&  6.02	&  7.09 \\
    \zestxml                  & 10.06	&  8.11	&  6.60	&  5.33	& 11.49	& 14.74	& 19.57	& 40.46 \\
    \sbert                    & 18.47	& 14.19	& 11.29	& 10.82	& 21.55	& 26.77	& 33.92	& 57.02 \\
    \mpnet			          & 18.59	& 13.99	& 11.08	& 10.89	& 21.12	& 26.10	& 32.82	& 54.70 \\
	\ours (ours)              & \bf20.99	& \bf15.57	& \bf12.26	& \bf12.59	& \bf23.94	& \bf29.41	& \bf36.78	& \bf59.81 \\
    \midrule
    \midrule
    \multicolumn{9}{c}{LF-Wikipedia-500K} \\
    \midrule
    \xrl	                  & 11.80	&  5.30	&  3.39	&  2.76	&  3.47	&  3.65	&  3.82	&  4.09 \\
    \astec                    & 11.23	&  5.27	&  3.48	&  5.46	&  7.47	&  8.16	&  8.90	& 10.35 \\
    \siamesexml               & 12.44	&  5.69	&  3.79	&  6.05	&  7.98	&  8.62	&  9.22	& 10.40 \\
    \zestxml                  & 27.31	& 17.31	& 13.09 &  8.28	& 15.13	& 18.64	& 23.30	& 36.50 \\
    \sbert	                  & 41.06	& 26.35	& 20.25	& 14.17	& 25.34	& 31.32	& 39.77	& 66.24 \\
    \mpnet			          & 42.81	& 28.07	& 21.66	& 14.67	& 26.81	& 33.24	& 42.28	& 67.76 \\
	\ours (ours)              & \bf47.25	& \bf30.57	& \bf23.54	& \bf16.20	& \bf29.01	& \bf35.81	& \bf45.13	& \bf71.35 \\
    \midrule
    \midrule
    \multicolumn{9}{c}{LF-Amazon-1M} \\
    \midrule
    \xrl	                  &  2.11 &  0.84 &  0.53 &  1.45 &  1.74 &  1.81 &  1.88 &  2.04 \\
    \astec                    &  2.22 &  2.56 &  0.71 &  1.54 &  1.91 &  2.03 &  2.16 &  2.41 \\
    \siamesexml               &  2.60 &  3.01 &  1.06 &  1.81 &  2.20 &  2.30 &  2.41 &  2.60 \\
    \zestxml                  &  7.17 &  8.35 &  6.36 &  5.18 & 17.49 & 21.88 & 26.80 & 36.51 \\
    \sbert                 	  &  8.89 & 10.02 &  7.93 &  7.00 & 21.58 & 27.35 & 33.98 & 54.28 \\
    \mpnet			          &  9.25 & 10.41 &  8.00 &  7.11 & 21.87 & 27.64 & 34.61 & 54.72 \\
	\ours (ours)              & \textbf{10.60} & \bf11.47 & \bf 8.80 &  \bf7.89 & \bf24.14 & \bf30.44 & \bf37.95 & \bf58.45 \\
    \bottomrule
\end{tabular}
\end{table}

Our proposed \ours outperforms all other baselines significantly,
	including variants of Siamese-Transformer models (e.g., \sbert, \mpnet) and
	major competitive \xmc methods (e.g., \xrl, \astec and \siamesexml), on all four datasets.
Note that \siamesexml is the state-of-the-art \xmc method under the full supervision setup of \xmc.
	Here, we again witness that existing \xmc methods heavily rely on the supervision level
	as well as the full-coverage of label space for test set.
\ours, in contrast, still performs robustly under the \fsxmc setup, 
which enjoy larger applicability to emerging domains with many cold-start labels.

Crucially, even \zestxml tailored to address the challenging scenario of unseen labels cannot match the performance of \ours.
In particular, when focusing on the few-shot scenario with only 1\% sampled labels,
	\ours achieves 18.74\% in P@1, improving the performance of \astec with 0.94\% and \zestxml with 10.10\% significantly.
Besides, \ours outperforms all other Sentence-BERT counterparts, validating the effectiveness of our pre-training procedure.
As to fine-tuning on the subset with 5\% labels, performance of all compared methods are improved as expected with more supervision.
The relative rank among these methods remains the same, with \ours still performing the best in terms of precision and recall on all four datasets.

\subsection{Ablation Study}
\label{sec:ablation}
In this part, we conduct an ablation study to investigate each component in our pre-training procedure,
including multi-scale adaptive clustering, label regularization, and self-training with pseudo positive pairs constructed from the encoder or TF-IDF.
We add a component once a time on LF-Amazon-131K to observe its independent influence on the model performance.
Table \ref{tab:ablation} presents detailed performance on seven different configurations.

For two techniques multi-scale adaptive clustering and label regularization during the Stage I, they can improve the performance of the encoder separately,
as shown in the performance gain of the index 2 and 3 over the index 1.
When combined, they can further improve the accuracy of the model, from 8.90\% to 10.65\% in $P@5$ and from 47.40\% to 51.45\% in $R@100$.
As to the second stage, we explore the impact of self-training with pseudo positive pairs either from the encoder itself or TF-IDF.
We can see from Table \ref{tab:ablation} that pairs from both $\gE$ and TF-IDF contribute to the improvement in precision and recall gain over the index 5.
It further validates that the encoder and TF-IDF provides complementary perspective when constructing pseudo positive pairs.
\begin{table}[!ht]
  \caption{Ablation study on LF-Amazon-131K.}
  \label{tab:ablation}
  \begin{threeparttable}[t]
  \centering
  \begin{tabular}{c|cccc@{ }|cccccccc}
  \toprule
  \multirow{2}{*} { Index} & \multicolumn{4}{c|}{ Ablation Configuration } &  \multicolumn{3}{c}{Precision}  & \multicolumn{5}{c}{ Recall} \\
      & MAC \tnote{*}   & LR \tnote{*} & $\mathcal{E}$ \tnote{$\dagger$} & TFIDF \tnote{$\dagger$} & @1 & @3 & @5 & @1 & @3 & @5 & @10 & @100 \\
  \midrule
  1   & No     & No     &   No    & No   & 13.82 & 11.41 &  8.90 &  7.76 & 18.09 & 22.80 & 28.94 & 47.40\\
  2   & Yes    & No     &   No    & No   & 15.79 & 13.16 & 10.22 &  8.85 & 20.90 & 26.27 & 32.61 & 49.83    \\
  3   & No     & Yes    &   No    & No   & 16.02 & 13.29 & 10.28 &  9.04 & 21.27 & 26.51 & 32.97 & 50.34    \\
  4   & Yes    & Yes    &   No    & No   & 16.37 & 13.71 & 10.65 &  9.29 & 21.63 & 27.03 & 33.93 & 51.45    \\
  \midrule
  5   & Yes    & Yes    &   Yes   & No   & 17.01 & 14.75 & 11.41 &  9.72 & 23.33 & 29.04 & 35.20 & 53.55   \\
  6   & Yes    & Yes    &   No    & Yes  & 16.51 & 14.12 & 10.92 &  9.52 & 22.43 & 28.02 & 34.64 & 52.78    \\
  7   & Yes    & Yes    &   Yes   & Yes  & 18.13 & 15.42 & 11.93 & 10.35 & 24.45 & 30.43 & 37.28 & 54.99   \\
  \bottomrule
  \end{tabular}
  \begin{tablenotes}
    \item[*] MAC represents adaptive clustering while LR stands for label regularization.
    \item[$\dagger$] Pseudo positive pairs are constructed from $\mathcal{E}$ or TFIDF.
  \end{tablenotes}
  \end{threeparttable}
\end{table}


\section{Conclusions}
This paper is the first to investigate the problem of Extreme zero-shot \xmc without any supervision.
We develop a two-stage pre-training procedure \ours to train a Sentence-BERT style encoder on pseudo (context, title) pairs constructed from raw text.
We demonstrate that techniques including multi-scale adaptive clustering, label regularization and self-training contribute to the performance gain of the pre-trained encoder.
In particular, \ours outperforms all unsupervised baselines significantly when there are no (instance, label) pairs provided. 
It also offers leading accuracy in both precision and recall after fine-tuning on a limited number of paired data, compared with state-of-the art extreme classifiers.
A future direction could be adding a ranker model after the encoder to improve performance on head labels.

\bibliography{ref.bib}
\appendix
\section{Additional Experiments on \fsxmc}
In this section, we present additional experimental results for the setting of \fsxmc on LF-Amazon-131K and LF-WikiSeeAlso-320K.
Instead of sampling a few-shot subset by the label coverage ratio, we turn to sampling based on the pair ratio.
Specifically, suppose a training set $\gD_\text{train}=\{(x_i, y_i)\}$ has $|\gD_\text{train}|$ positive pairs.
Each time we randomly sample a small ratio of $\delta$~(1\% or 5\% in our paper) pairs from the total set to constitute the few-shot subset.
Then each subset has $\delta|\gD_\text{train}|$ pairs for fine-tuning.
Detailed results of $\delta=1\%$ and $\delta=5\%$ are presented in Table~\ref{tab:fs-pair-1} and \ref{tab:fs-pair-5} respectively.
\ours is still the best-performing method and outperforms all other baselines significantly in precision and recall.
\begin{table}[!ht]
	\caption{Results of \fsxmc where the training subset covers 1\% positive pairs from the whole set.}
	\label{tab:fs-pair-1}
	\centering
  \begin{tabular}{l|rrr|rrrrr}
	  \toprule
  \multirow{2}{*}{Method}   & \multicolumn{3}{c|}{Precision} &\multicolumn{5}{c}{Recall} \\
  & @1 & @5 & @10 & @1 & @3 & @5 & @10 & @100 \\
	  \midrule
	  \multicolumn{9}{c}{LF-Amazon-131K} \\
	  \midrule
	  \xrl	                  &  5.37	&  2.66	&  1.68	&  2.81	&  3.92	&  4.09	&  4.26	&  4.99 \\
	  \astec                    &  3.29	&  2.04	&  1.41	&  1.93 &  3.33	&  3.77	&  4.06	&  5.06 \\
	  \siamesexml               & 7.14   &  3.74	&  2.41	&  4.22	& 6.17	& 6.55	& 6.95	& 8.09 \\
	  \zestxml                  & 12.91   &  11.31	&  8.91	&  7.20	& 17.69	& 22.51	& 28.27	& 42.40 \\
	  \sbert                    & 15.08	&  11.81	&  9.06	&  8.38	& 18.42	& 22.89	& 28.62	& 46.38 \\
	  \mpnet		              & 15.26	& 12.30	&  9.42	&  8.56	& 19.35	& 23.98	& 29.91	& 48.06 \\
	  \ours (ours)              & \bf18.92	& \bf16.17	& \bf12.62	& \bf10.98	& \bf25.64	& \bf32.16	& \bf39.46	& \bf58.24 \\
	  \midrule
	  \midrule
	  \multicolumn{9}{c}{LF-WikiSeeAlso-320K} \\
	  \midrule
	  \xrl	                  &  6.97	&  3.43 &  2.31	&  3.74	&  5.02	&  5.44	&  5.84	&  6.87 \\
	  \astec                    &  5.58	&  3.35	&  2.48	&  3.22 &  5.43	&  6.51	&  7.95	&  11.76 \\
	  \siamesexml               &  9.87	&  5.22	&  3.59	&  5.84	&  8.57	&  9.53	&  10.60	&  13.04 \\
	  \zestxml                  &  10.40	&  8.18	&  6.49	&  5.57 &  11.65	& 14.52	& 18.81	& 33.20 \\
	  \sbert                    & 18.85	& 14.23	& 11.22	&  11.16	& 21.77	& 26.94	& 33.78	& 55.88 \\
	  \mpnet			          & 18.04	& 13.27	&  10.44 &  10.51	& 19.99	& 24.62	& 30.86	& 52.52 \\
	  \ours (ours)              & \bf20.49	& \bf15.50	& \bf12.24	& \bf12.34	& \bf23.88	& \bf29.43	& \bf36.76	& \bf59.82 \\
	  \bottomrule
  \end{tabular}
  \end{table}
  \begin{table}[!ht]
	\caption{Results of \fsxmc where the training subset covers 5\% positive pairs from the whole set.}
  \label{tab:fs-pair-5}
	\centering
	\begin{tabular}{l|rrr|rrrrr}
	  \toprule
		 \multirow{2}{*}{Method}   & \multicolumn{3}{c|}{Precision} &\multicolumn{5}{c}{Recall} \\
		& @1 & @5 & @10 & @1 & @3 & @5 & @10 & @100 \\
	  \midrule
	  \multicolumn{9}{c}{LF-Amazon-131K} \\
	  \midrule
	  \xrl	                  &  11.20	&5.82	&3.80	&5.98	&8.56&	9.18&	9.80&	12.79	  \\
	  \astec                    &  10.71&	6.50	&4.52	&6.12	&10.23&	11.67&	13.35&	18.15	  \\
	  \siamesexml               &  11.88&	8.72	&5.93	&8.50	&13.68&	15.23	&16.80&	20.28	  \\
	  \zestxml                  & 12.86	&11.28	&8.91	&7.10&	17.62	&22.43	&28.42&	49.41\\
	  \sbert	                  & 16.94	&13.59&	10.52	&9.55&	21.23&	26.55&	33.14&	51.81 \\
	  \mpnet			          & 17.48	&13.58	&10.61	&9.95&	21.38	&26.83&	33.60&	52.31\\
	  \ours (ours)              & \bf 19.75 & 	\bf16.45 &	\bf12.87&	\bf11.18	&\bf25.99 & \bf32.70 & \bf40.38	& \bf59.82\\
	  \midrule
	  \midrule
	  \multicolumn{9}{c}{LF-WikiSeeAlso-320K} \\
	  \midrule
	  \xrl                 	  &  13.13&	6.88&	4.70&	7.00&	9.64&	10.54&	11.49&	14.20\\
	  \astec                    &  15.61&	8.73&	6.23&	8.77&	13.17&	15.02&	17.36&	24.30\\
	  \siamesexml               &  16.51&	9.68	&6.96	&9.40&	14.78	&16.97&	19.48&	25.26\\
	  \zestxml                  & 17.68	&8.51&	6.85	&10.63&	12.01&	15.20	&20.08&	43.10\\
	  \sbert                    & 20.12&	15.01&	11.87&	12.05&	23.01&	28.40&	35.52&	58.41\\
	  \mpnet			          & 19.88	&14.90	&11.76	&11.85&	22.75&	27.96&	35.03&	57.26\\
	  \ours (ours)              & \bf21.80& \bf16.61& \bf13.12 & \bf13.27& \bf25.74 & \bf31.59& \bf39.25& \bf62.13\\
	  \bottomrule
  \end{tabular}
  \end{table}

\end{document}